\documentclass[10pt,twocolumn,letterpaper]{article}

\usepackage[final]{cvpr}      
\usepackage{colortbl}
\usepackage{wrapfig}
\usepackage{adjustbox}
\usepackage{graphicx}
\usepackage{colortbl}
\usepackage{multirow}
\usepackage{algorithm}
\usepackage[noend]{algorithmic}
\usepackage{listings}
\definecolor{cvprblue}{rgb}{0.21,0.49,0.74}
\usepackage[pagebackref,breaklinks,colorlinks,allcolors=cvprblue]{hyperref}

\def\paperID{****} 
\def\confName{CVPR}
\def\confYear{2025}

\title{Spatiotemporal Predictive Pre-training
for Robotic Motor Control}

\author{Jiange Yang $^{1}$ \quad Bei Liu$^{2}$ \quad Jianlong Fu$^{2}$ \\
Bocheng Pan$^{3}$ \quad Gangshan Wu$^{1}$ \quad Limin Wang$^{1}$\thanks{Corresponding author}\\
$^1$Nanjing University, $^2$Microsoft Research, \\
$^3$Institute of Microelectronics of the Chinese Academy of Sciences
}

\begin{document}
\maketitle

\begin{abstract}
  Robotic motor control necessitates the ability to predict the dynamics of environments and interaction objects. However, advanced self-supervised pre-trained visual representations in robotic motor control, leveraging large-scale egocentric videos, often focus solely on learning the static content features. This neglects the crucial temporal motion clues in human video, which implicitly contain key knowledge about interacting and manipulating with the environments and objects. In this paper, we present a simple yet effective robotic motor control visual pre-training framework that jointly performs spatiotemporal prediction with dual decoders, utilizing large-scale video data, termed as \textbf{STP}. STP adheres to two key designs in a multi-task learning manner. First, we perform spatial prediction on the masked current frame for learning content features. Second, we utilize the future frame with an extremely high masking ratio as a condition, based on the masked current frame, to conduct temporal prediction for capturing motion features. The asymmetric masking and decoupled dual decoders ensure that our image representation focusing on motion information while capturing spatial details. Extensive simulation and real-world experiments demonstrate the effectiveness and generalization abilities of STP, especially in generalizing to unseen environments with more distractors. Additionally, further post-pre-training and hybrid pre-training unleash its generality and data efficiency. Our code and weights will be released for further applications.
\end{abstract}

\vspace{-6mm}



\section{Introduction}
\label{sec:intro}

\vspace{-0.5mm}


In NLP and CV, adapting pre-trained foundation models from large-scale data to various downstream tasks has seen great success. For example, pre-trained visual representations using self-supervised~\cite{mae,moco3,dinov2,jepa,videomae2} or weakly-supervised~\cite{clip,eva,egoclip} methods exhibit strong generalization ability. However, in robot learning, due to data scarcity and homogeneity, some groundbreaking methods~\cite{scratch1,scratch12} resort to training from scratch only using in-domain data. Recently, inspired by the success of transfer learning in CV, many works~\cite{robotclip,mvp,r3m,vip,vc1,unbiased_dataset} have explored developing a pre-trained visual representation (PVR) using large-scale out-of-domain data for various robotic motor control tasks. Currently, one successful paradigm~\cite{mvp,mvp2,vc1,unbiased_dataset} is to use large-scale egocentric video datasets~\cite{ego4d} and train vanilla vision transformers (ViT)~\cite{vit} based on MAE~\cite{mae}, which exhibits excellent learning efficiency and generalization ability for learning policy from raw pixel. Among them, the Ego4D~\cite{ego4d} dataset offers numerous first-person human-object interaction scenes and good motion clues. We argue that although learning static spatial structure priors from task-relevant pre-training data sources is crucial, designing a more relevant self-supervised proxy task for motor control should not be overlooked. Therefore, in this paper, we aim to develop a more relevant self-supervised proxy task for robotic motor control representation learning.







Robotic motor control typically requires fine-grained spatial localization and relatively dense semantics. With its ability to effectively capture low-level geometry and space structure, MAE~\cite{mae} pre-training excels at this task. However, is dense spatial content sufficient for robotic motor control? Some neuroscientific studies~\cite{neural,neural2,neural3} suggest the brain's different areas show specialization. Some are dedicated to processing the temporal object motion, while others focus on static spatial details. Their combination results in subjective pattern perception. Inspired by this finding, we hypothesize that an effective robotic motor control pre-training proxy task should require joint learning of spatial content features and temporal motion features. However, current methods~\cite{mvp,vc1,unbiased_dataset} use MAE pre-training with image frames from human videos, capturing only static content features. They overlook the temporal motion clues in human videos, which implicitly contain key knowledge about interaction with environment and manipulation of objects. Therefore, we aim to bridge this gap by incorporating these motion clues into our proxy task.



 
Based on the analysis above, the most critical challenge is the absence of action annotations in human video for modeling object motion. To self-supervisedly model interaction and manipulation actions from actionless video, we propose to implicitly capture them by predicting future frame pixels based on current frame. However, predicting the future frame without any conditions could contain high uncertainty and be extremely difficult. Therefore, we propose to use the future frame with an extremely high masking ratio as a prompt condition, specifically 95\%, which serves to reveal some behavior and dynamic priors, i.e. what to do and how to do it. However, directly and simply executing temporal prediction could lead the model to overlook static spatial details, and it is also not efficient enough. Therefore, another technical contribution of STP is to jointly perform spatial prediction by masking the current frame with 75\% masking ratio. In summary, we present STP, a \textbf{multi-task self-supervised} image pre-training framework through spatiotemporal prediction. STP asymmetrically mask the current and future frame from a video clip, using a spatial decoder to conduct spatial prediction for content learning and a temporal decoder to conduct temporal prediction for motion learning. The \textit{asymmetric masking and decoupled dual decoders} ensure that our pre-trained image representation focusing on motion information while capturing spatial details. Additionally, our image representation pre-training does not rely on any extra supervision, specific policy architecture or application scenarios, and intricate modules, making our method more general, flexible and scalable.

We employ a data-efficient few-shot behavior cloning paradigm as our evaluation scheme. our primary evaluation scheme involves freezing the image encoder during policy training. Additionally, considering that fine-tuning ViT with few demonstrations might lead to overfitting and masked modeling exhibits excellent data efficiency~\cite{videomae,comae,occlu} in domain-in data, we also further follow the post-pre-training~\cite{beit,videomae2,vc1} adaptation paradigm to perform STP pre-training with task-specific data to achieve better results. Finally, we conduct extensive simulation and real-world experiments to demonstrate the effectiveness and generalization capabilities of STP. These evaluations encompass various embodiments (single-arm, dual-arm, dexterous hands, and humanoid), different object types (deformable and transparent objects), and different settings (single-task, multi-task and generalized setup). Moreover, we also yield some insightful observations.

We make the following four contributions: \textcolor{blue}{\textbf{(1)}} We present STP, a \textit{self-supervised} image pre-training framework for robotic motor control, which jointly conducts spatiotemporal prediction with \textit{asymmetric masking and decoupled dual decoders} for content and motion features learning. \textcolor{blue}{\textbf{(2)}} We further expand STP by performing hybrid pre-training with ImageNet-MAE and post-pre-training with task-specific data, unleashing its \textit{generality} and \textit{data efficiency}. \textcolor{blue}{\textbf{(3)}} Our extensive simulation and real-world experiments demonstrate the effectiveness and generalization capabilities of STP, especially in generalizing to \textit{unseen environments with more distractors}. \textcolor{blue}{\textbf{(4)}} Our experiments yield some insightful observations. As for temporal prediction, language does not significantly enhance performance in single-task evaluation. Instead, \textit{single-modality self-supervised paradigm} achieves the best results. Moreover, in the few-shot BC setting, naively scaling up model size does not necessarily lead to improved outcomes. Finally, compared with fine-tuning, incorporating \textit{more diverse} data and \textit{domain-in} data into the pre-training can further enhance downstream performance.

\vspace{-3.0mm}

\begin{figure*}[htbp]
\centering
\includegraphics[width=0.975\textwidth]{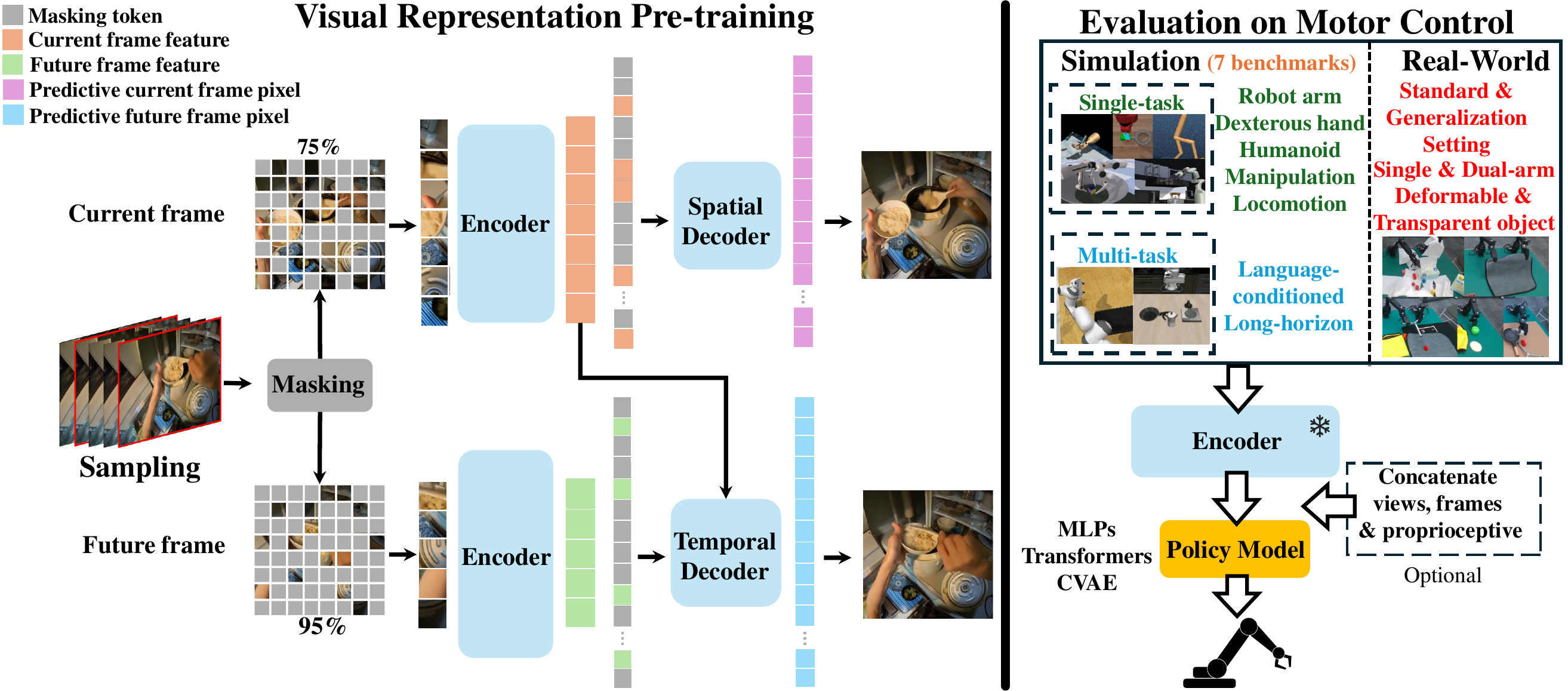} 
\label{pipelines}
\vspace{-0.5mm}
\caption{\textbf{STP framework}. \textbf{Left:} During pre-training, we sample the current frame and the future frame from the video clip, and carry out spatiotemporal predictive pre-training. \textbf{Right:} During extensive downstream motor control tasks evaluation, we freeze the pre-trained encoder to extract visual state representations and discard the decoders.}
\label{pipelines}
\vspace{-5mm}
\end{figure*}


\section{Related Work}
\vspace{-1.0mm}


\textbf{Pre-trained Visual Representation Learning.} Large-scale visual representation pre-training is increasingly enhancing computer vision, with supervised learning methods primarily focusing on image recognition~\cite{resnet,deit} from ImageNett~\cite{imagenet} and multi-modal alignment from image-text pairs~\cite{clip}. Self-supervised learning methods, currently gaining popularity, are mainly divided into two categories: one that uses contrastive learning~\cite{moco,moco3,simclr} or joint-embedding~\cite{dino} to learn view-invariance, and another that employs masked modeling~\cite{beit,mae,simmim,maskfeat,data2vec,jepa} to predict invisible spatial parts. Additionally, some approaches~\cite{ibot,dinov2,mc-jepa} propose combining various optimization objectives in a multi-task learning framework. Recently pre-trained visual representation learning for robotic motor control have bee rapidly developing~\cite{robotclip,r3m,mvp,mvp2,vip,liv,language_driven,vc1,unbiased_dataset,decisionnce,mpi,theia,spa}. In contrast, our STP conducts pre-training on real-world video data and plain image ViT structure. It is entirely self-supervised, requiring no additional language, action, depth and teacher supervision, specific policy architecture and application scenarios, as well as intricate modules. Finally, we conduct extensive simulation and real-world experiments to demonstrate the effectiveness and generalization capabilities of STP across various embodiments, different object types, and different settings, yielding some insightful observations.

\noindent \textbf{Temporal Predictive Learning.} Early works once explored representation learning through future prediction~\cite{lecun,videoDPC,LSTM,audiocpc}.  Recently TrackMAE~\cite{mixformer_j} and SiamMAE~\cite{siammae} improved object tracking and segmentation by predicting masked future frames from unmasked current frames, enhancing the capture of temporal correspondence. In robot learning, temporal prediction primarily serves as a transition dynamic model such as World Models~\cite{SWM,ava,mwm}. GR-2~\cite{gr1,gr2} and PVDD~\cite{xuelong} conducts language-conditioned video prediction for policy pre-training in frozen representation space. ~\cite{plex,robocat} predict the future states using goal image in robot data. ~\cite{dynamics_aware} proposed dynamics-aware representation learning, and~\cite{smart,sensormotor} employed forward dynamics for self-supervised policy pre-training. Some works explored to train video prediction models and utilize visual foresight~\cite{maskvit, manipulate_by_seeing}, inverse dynamics models~\cite{diffusion_inverse}, goal-conditioned policy learning~\cite{VLP}, and geometry estimation~\cite{flow} for motor control, respectively. Different from these works, we aim to utilize the large-scale egocentric video data and employ masked spatiotemporal prediction as a self-supervised proxy task to pre-train a plain ViT image representation. Therefore, our approach requires no additional action or language annotations, nor does it rely on specific policy architecture, application scenarios and intricate modules, making it more general, flexible and scalable. Additionally, unlike VideoMAE~\cite{videomae,videomae2}, which operates reconstruction on video architecture, our STP employs decoupled and asymmetric masking and decoders to separately capture image-level content and motion features.






\noindent \textbf{Vision-based Robot Learning.} Vision-based robot learning is pivotal in robotics. Recently some works study model design~\cite{Explore,openworld,dual}, observation spaces~\cite{point,3d_diff}, policy algorithms~\cite{equal,act,diff}, sim-to-real ~\cite{sim2real}, adapters~\cite{lossless,spawnnet}, learning-from-scratch~\cite{lfs}, and affordance~\cite{affordances,f_affordance,abc,task_ada} in visuo-motor learning. Other related works~\cite{dexmv,human,mimicplay,xskill,giving,atm} attempt to learn skill knowledge from small-scale videos. Additionally, language-conditioned vision robot learning is increasingly recognized. Some works scale multimodal robot data~\cite{bcz,rt1,scale,bridge,rh20t,rtx,octo,openvla,kaiming} or introduce Internet data \cite{moo,pave,rt2,roboflamingo,alphablock,poco,embodiedgpt} for end-to-end robot learning. In our study, we pre-train a off-the-shelf ViT image representation from large-scale egocentric video datasets for extensive robotic motor control tasks. Our image representation is simple and general for different policy learning paradigms of motor control.

\vspace{-2.0mm}

\section{Method}
\vspace{-2mm}

In this section, we describe our method in details. First, we give an overview of our spatiotemporal predictive pretraining (STP) framework. Then, we give a technical description on our core components during pre-training: the masked image encoder and dual decoders scheme. Finally, we describe how to adapt our pre-trained encoder to downstream robotic motor control tasks.

\vspace{-2mm}

\subsection{Overiew of STP}

As illustrated in Fig.~\ref{pipelines}, our STP aims to pre-train an image encoder for robotic motor control from video datasets. This pre-trained image encoder is subsequently frozen and directly transferred to solve motor control tasks. Here, we highlight that our STP differs from some video pre-training works~\cite{videomae,language_driven,mpi} in that it uses a plain ViT encoder to encode individual image frames without extra cross-frame attention operation, while employing decoupled and asymmetric masking and decoders for spatial and temporal prediction, without additional intricate modules.




 \vspace{-2mm}

\subsection{Masked Image Encoder}


We first introduce the pipeline of our image encoder. Our image encoder processes image frames using a vanilla vision transformer~\cite{vit}. Given a image $\mathbf{I} \in \mathbb{R}^{C \times H \times W}$, we initially process it by the patch embedding layer to obtain its token sequences $\mathbf{T}$, where $\mathbf{T} = \{P_i\}_{i=1}^{N}$ and $N$ is the the total token number, (e.g., N = 196 for a 224 × 224 image with a patch size of 16 × 16). Then we add the fixed 2D sine-cosine positional embeddings for all tokens. Following this, we mask and remove a part of tokens, according to a randomly generated masking map $\mathbb{M}(\rho)$, where $\rho$ is the masking ratio. The encoder applies several transformer blocks (consisting of a global self-attention layer and a FFN layer) on all unmasked tokens: $\mathbf{Z} = \Phi_{enc}(\mathbf{T}^{u})$, where $\mathbf{T}^u = \{T_i\}_{i \in (1-\mathbb{M}(\rho))}$. During this process, a [CLS] token is added at the beginning.

\begin{figure*}[htbp]
\centering
\includegraphics[width=0.90\textwidth]{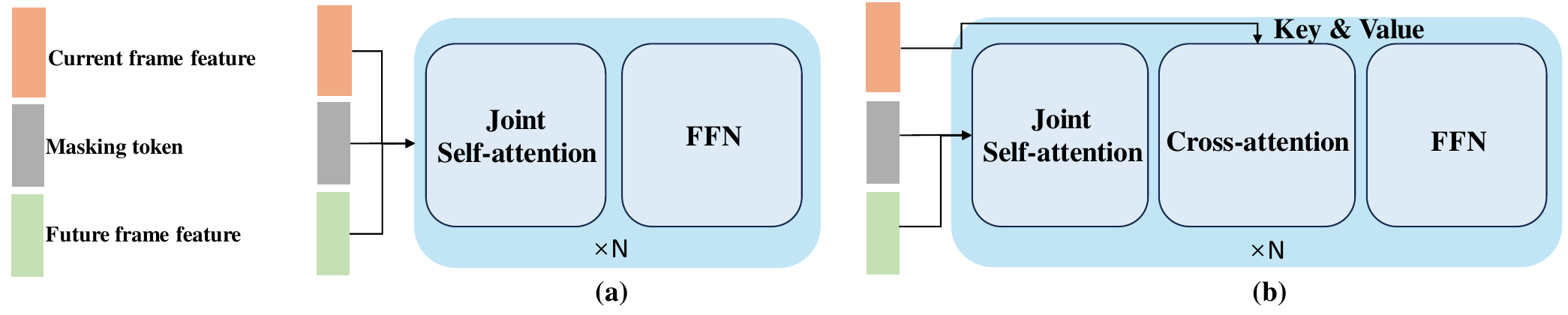} 
\label{decord}
\vspace{-0.5mm}
\caption{Temporal decoder design. \textbf{(a)} Standard joint-self architecture. \textbf{(b)} The self-cross architecture.}
\label{decord}
\vspace{-4mm}
\end{figure*}



Then we describe our encoding process during pre-training. We randomly sample two frames from a video clip based on an interval: the current frame $\mathbf{{I}_{c}}$ and the future frame $\mathbf{{I}}_{f}$. Following the above pipeline, we randomly generate two asymmetric masking maps for the current frame and the future frame, denoted as $\mathbb{M}_c = \mathcal{M}_c(\rho^c)$ and $\mathbb{M}_f = \mathcal{M}_f(\rho^f)$, respectively. Each of these maps has a different masking ratio. We then use these maps to separately process the two frames and obtain their features, $\mathbf{{Z}}_{c}$ and $\mathbf{{Z}}_{f}$.  As analyzed above, our STP aims to jointly learn content and motion features by spatiotemporal predictive learning. For content feature learning, we follow MAE~\cite{mae}, masking a portion of the current frame based on $\mathbb{M}_c$, with $\rho^c = 75\%$, and predict the masked parts during the decoding process. This encourages the model to learn spatial and geometric structure priors from the current frame data through spatial reasoning. For motion feature learning,  we establish an objective to predict the future frame based on the masked current frame. However, predicting the future frame without any conditions could be meaningless and extremely challenging. Therefore, we use the future frame with an extremely high masking ratio as a condition, specifically $\rho^f = 95\%$, which reveals some behavior and dynamic priors. In the experiments section, we will further discuss different condition schemes, including language narration and the combination between them. In summary, our encoding process during pre-training can be formally described as follows:

\begin{equation}
\label{eq:c(5)}
     \begin{cases}
\mathbf{Z}_c= \Phi_{enc} (\mathbf{I}_{c}, \mathbb{M}_c),
 \\
\mathbf{Z}_f= \Phi_{enc} (\mathbf{I}_{f}, \mathbb{M}_f).
\end{cases}
\end{equation}



\subsection{Dual Decoders}


To jointly capture static content and object motion features for better spatiotemporal understanding, our STP present a dual decoders scheme to predict both the pixel of current and future frame simultaneously in a multi-task learning manner. As shown in Fig.~\ref{pipelines}, our dual decoder scheme includes a spatial decoder $\Phi_{dec\_s}$ for spatial prediction and a temporal decoder $\Phi_{dec\_t}$ for temporal prediction. We firstly give a technical description on them, respectively. Then we describe how we combine them into our final method.

\noindent \textbf{Spatial Decoder.} To capture static content features, our spatial decoder is solely utilized for processing the current frame visual feature. Specifically, after obtaining the masked current frame visual feature $\mathbf{{Z}}_{c}$, we concatenate it with some learnable masking tokens, leading to the formation of $\mathbf{Z}_{c}^{d} = \mathbf{Z}_{c} \cup \{\mathbf{M}_i\}_{i \in \mathbb{M}_c}$, where $\mathbb{M}_c$ is the current frame masking map. Then, each of these tokens further adds a corresponding positional embedding.  Subsequently, $\mathbf{Z}_{c}^{d}$ undergoes decoding in the decoder and is continuously updated. The architecture of the spatial decoder block aligns with the standard transformer encoder block, comprised of a global self-attention layer and a FFN layer. Finally, with the deocoded token sequence $\mathbf{Z}_{c}^{d}$, our spatial decoder predicts the invisible tokens of the current frame $\hat{\mathbf{I}_{c}^{d}}$, operating under the current frame masking map $\mathbb{M}_c$.



\noindent \textbf{Temporal Decoder.} To capture motion features, our temporal decoder jointly processes the current and the future frame which serves as the temporal prediction condition. To elaborate, we firstly obtain the current frame feature $\mathbf{{Z}}_{c}$ and the future frame feature $\mathbf{{Z}}_{f}$. We then concatenate $\mathbf{{Z}}_{f}$ with the masking tokens that have the positional embedding added, resulting in $\mathbf{{Z}}_{f}^{d}$. Following this, $\mathbf{{Z}}_{f}^{d}$ and $\mathbf{{Z}}_{c}$ interact within the temporal decoder for decoding. The design of temporal decoder block is in alignment with the standard transformer decoder~\cite{transformer}, consisting of a self-attention layer, a cross-attention layer, and a FFN layer, as shown in Fig.~\ref{decord} (b). During decoding, the self-attention layer and FFN are solely used to process $\mathbf{{Z}}_{f}^{d}$. For the cross-attention layer, like some cross-interaction works~\cite{multimae, mixformer_j, siammae}, $\mathbf{{Z}}_{f}^{d}$ is continuously updated as the query, while $\mathbf{{Z}}_{c}$, acting as the key and value, is kept constant. Compared to standard architecture, it ensures that the current frame representation space will not be updated in the temporal decoder and cannot see the future frame. The current frame specifically acts as condition for  temporal correlation and prediction. This asymmetric interact architecture not only achieves more efficient training but also produces better results. Finally, with the decoded token sequence $\mathbf{Z}_{f}^{d}$, our temporal decoder predicts the invisible tokens of the future frame $\hat{\mathbf{I}_{f}^{d}}$, operating under the future frame masking map $\mathbb{M}_f$.

\noindent \textbf{Multi-task Predictive Learning.} As mentioned above, STP jointly conducts spatiotemporal prediction by asymmetric masking ratio and dual decoders scheme, the whole decoding pipeline can be formally described as follows:

\begin{equation}
\label{eq:c(5)}
     \begin{cases}
\hat{\mathbf{I}}_{c}^{d}= \Phi_{dec\_s} (\mathbf{Z}_{c}^{d}),
 \\\hat{\mathbf{I}}_{f}^{d}= \Phi_{dec\_t}
 (\mathbf{{Z}}_{c}, \mathbf{Z}_{f}^{d}).
\end{cases}
\end{equation}

Our loss function is the mean squared error (MSE) loss between the normalized masked pixels and the predicted pixels. So our loss function $\ell$ is as follows: 
\begin{equation}
\label{eq:c(6)}
\ell = \mathrm{MSE}(\hat{\mathbf{I}}_c, \mathbf{I}_c) + \mathrm{MSE}(\hat{\mathbf{I}}_f, \mathbf{I}_f).
\end{equation}

\subsection{Downstream Policy Learning}



To enable data and computation efficiency, we adopt the few-shot behavior cloning paradigm and keep the image encoder frozen. Concretely, given some offline expert demonstrations $\mathcal{S} = \{\tau_1, . . . , \tau_n\}$, where each $\tau_i$ is a trajectory of
robot observations and actions, denoted as $\tau_i = [(o_0, a_0), \ldots, (o_T, a_T)]$. Based on the $\mathcal{S}$, we train a  policy model, $\pi_{\theta}{(a|\mathcal{C}(\Phi_{enc}(o)), l)}$,
parameterized by $\theta$, which maps from visual representations to actions. Here, $l$ represents optional task instruction and $\mathcal{C}$ represents an optional concatenation operation that fuses multi-view and multi-frame visual features, along with the robot's proprioceptive state in the channel dimension. We optimize the $\pi_{\theta}$ through a standard behavior cloning MSE loss:

\vspace{-3mm}

\begin{equation} 
\min_{\theta} {\textstyle \sum_{  \boldsymbol{ (o,a)\sim\mathcal{S}}}}\mathrm{MSE}(a, \pi_{\theta}{(\mathcal{C}(\Phi_{enc}(o)), l)}).
\end{equation}



\vspace{-3mm}

\section{Experiments}

\subsection{Implementation on Pre-training}

We execute pre-training with data from EgoVLP~\cite{egoclip} for comprehensive ablation and fair comparison. It processes untrimmed videos of Ego4D and filters out that miss language narrations and belong to validation or test sets, resulting in a total of 3.8 million clips, called as Egoclip. 
In pre-training, we sample a frame pair from each clip for training. As for all experiments, we employ the standard ViT~\cite{vit} as backbone. We train ViT-B/16 and ViT-L/16 two versions. In the experiments section, unless otherwise specified, all models use ViT-B/16. Additionally, we maintain consistency with prior works~\cite{mvp,vc1}, directly using the [CLS] token as the global representation. The pre-training hyperparameters can be found in our appendix. 



\subsection{Implementation on Downstream Policy}







\textbf{Simulation Tasks.} In our simultion experiments, we select the union of manipulation and locomotion tasks from prior works~\cite{r3m,vc1} for single-task setup evaluation, encompassing 19 tasks across 5 simulated environments. These inclue Meta-World~\cite{metaworld}, Franka-Kitchen~\cite{franka}, Adroit~\cite{adroit}, DMControl~\cite{dmc}, and Trifinger~\cite{trifinger}. Additionally, we further strengthen our evaluation under the multi-task setup, which includes 20 tasks randomly selected from RLbench~\cite{rlbench} and 10 long-horizon tasks from LIBERO~\cite{libero}, referred to as LIBERO-LONG. More detailed simulation evaluation details can be found in our appendix.

\noindent \textbf{Real-World Tasks.} In our real-world experiments, we evaluate 5 tasks using a open-sourced
Low-Cost Robot Arm. This includes two single-arm tasks: pouring water and pushing a vegetable to the side of the cutting board, as well as three dual-arm tasks: pulling out tissues, folding towels, and picking up and passing a holder. These tasks cover the manipulation of transparent and deformable objects. For each task, we collect 50 noisy demonstrations for training and conduct 20 trials during evaluation. Specifically, we perform evaluation in both the standard environment and unseen environment with more distractor objects to test generalization. Our real-world evaluation setup and several evaluation demonstrations are shown in Fig.~\ref{reals}. Please see our appendix for more real-world setup details. 



\noindent \textbf{Evaluation Scheme.} For our single-task simulation benchmarks, following prior works~\cite{r3m,language_driven,vc1}, for each task, we learn a single policy $\pi$ which is structured as a MLPs network. For our multi-task simulation benchmarks, in Libero-Long~\cite{libero}, we employ the official transformer policy; for RLBench~\cite{rlbench}, we utilize the multi-view transformer from RVT-2~\cite{rvt} as the policy. In our real-world experiments, We learn an ACT~\cite{act} policy for each task. Here, in our simulation multi-task and real-world experiments, we run the tests three times and report the mean and standard deviation. In our simulation single-task experiments, we evaluate each task with different random seeds, which results in a total of \textbf{2,450} trials across five simulated environments. Different random seeds leads to different initial rendered images; therefore, we only report the mean from all trials. Additionally, we also run the comparison between STP and MAE three times to demonstrate the stability of our improvments, with details in our appendix.

\begin{figure*}[t]
\centering
\includegraphics[width=0.96\textwidth]{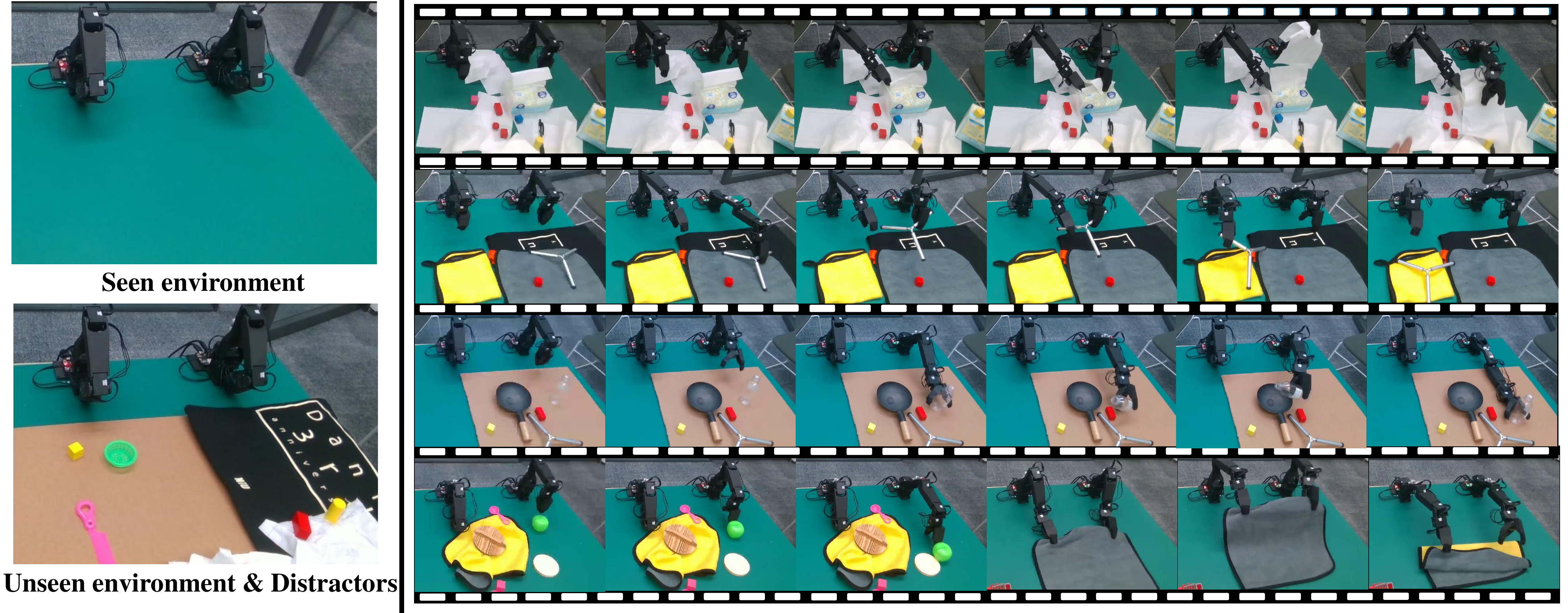} 
\label{reals}
\vspace{-0.5mm}
    \caption{The real-world experiments setup and five evaluation demonstrations. }
\label{reals}
\vspace{-4.0mm}
\end{figure*}




\subsection{Performance on Downstream Tasks}


In this section, we mainly analyze the performance of pre-trained image representations. Specifically, we first evaluate the following models: (1) public DINOv2~\cite{dinov2} that combines masked modeling with self-distillation on large-scale image datasets; (2) public CLIP~\cite{clip} that conducts contrastive learning on large-scale image-text pairs; (3) R3M-ViT trained based on Egoclip~\cite{egoclip}; (4) public VC-1~\cite{vc1}; (5) MAE trained based on Egoclip; (6) STP trained based on Egoclip. (7) STP that conducts hybrid pre-training with initialization using ImageNet-MAE~\cite{mae}. Among them, (1) and (2) achieve excellent performance on core visual understanding tasks using zero-shot or linear probing evaluation settings. (3) and (4) utilize egocentric videos for robotic motor control. (5), (6) and (7) are used for fair comparison and exploring the potential benefits of STP from more diverse image data, respectively. The simulation single-task experimental results are presented in Tab.~\ref{sota}. Next, we also evaluate and compare the adaptation results of representations to downstream motor control tasks. Specifically, we evaluate following settings: (a) The MAE pre-trained representation undergoes further MAE post-pre-training with task-specific data; (b) The STP pre-trained representation undergoes further STP post-pre-training with task-specific data; (c) The STP pre-trained representation undergoes end-to-end fine-tuning with task-specific data; (d) STP pre-training is performed directly using task-specific data. Finally, we also further conduct a fair comparison between (7) and (1)-(5) under the simulation multi-task setup and real-world setup, with the results shown in Tab.~\ref{sota2} and Tab.~\ref{sota33}, respectively. In summary, our extensive simulation and real-world experiments yield the following findings.


\noindent \textcolor{blue}{{\textbf{\textit{Finding 1:}}}} There is not a universal foundation model that performs optimally across all benchmarks. However, on the whole, the MAE (VC-1) method is superior due to its effective modeling of spatial geometry, especially for the fine-grained MetaWorld benchmark. Another intriguing observation is that MAE (VC-1) underperforms in the Franka-Kitchen and Adroit tasks. We believe that this could be due to its relatively weaker semantic representation.


\noindent  \textcolor{blue}{{\textbf{\textit{Finding 2:}}}} Our STP consistently outperforms our MAE baseline (the same pre-training data) and the public VC-1, achieving the best overall results. Specially, our STP outperforms MAE by 4.1 (59.6 → 63.7) in single-task simulation, and additionally benefits from a more diverse image data, improving by 0.5 (63.7 → 64.2). Moreover, STP achieves improvements of 6.1 (45.2 → 51.3) in LIBERO-LONG and 4.0 (37.3 → 41.3) in RLbench. This is attributed to that our STP not only captures static content features but also models motion information. We provide the visualization of the attention maps (model (5) and (6)) of several specific tasks in Fig.~\ref{viss}. The results indicate that, on top of effectively capturing spatial information, our method further encourages the model to focus on motion areas or objects, thereby providing a more \textit{sparse and compact} representation for downstream low-data BC paradigm. 

\noindent  \textcolor{blue}{{\textbf{\textit{Finding 3:}}}} End-to-end fine-tuning fails to yield the best results. This suggests that naively fine-tuning ViT-base could still lead to overfitting under few-shot BC scheme. Conversely, (a) and (b) achieve competitive adaptation results, with our STP achieving a 3.9 (72.5 → 76.4) improvement rate than MAE, further demonstrating the effectiveness and data efficiency of STP for in-domain data. In addition, the comparison between (a) and (d) also proves the effectiveness of pre-training with out-of-domain data.

\noindent  \textcolor{blue}{{\textbf{\textit{Finding 4:}}}} Simply expanding model capacity does not necessarily lead to performance gains. We also scale up STP to ViT-L/16, compared to VC1-L/16, and the results still demonstrate the superiority of STP. Among them, compared to ViT-B/16, ViT-L/16 brings a smaller performance improvement, which may be due to the task's performance saturation. However, the ViT-L/16 of STP does not show improvement in Meta-World, Trifinger and LIBERO-LONG. The larger visual backbone typically leads to a larger policy model, and we believe that there is a risk of overfitting during both policy training and backbone adaptation.


\noindent  \textcolor{blue}{{\textbf{\textit{Finding 5:}}}} Our STP demonstrates greater advantages in real-world environments, particularly in generalizing to unseen environments with more distractors. 
Specifically, in the standard environment and unseen environment with more distractors, our STP outperformed VC-1 by 7.3 (42.0 → 49.3) and 9.6 (36.7 → 46.3) respectively.

\begin{table*}[t]
\centering
\begin{tabular}{c|c|c|c|c|c|c|c}
\hline
        & Pre-training Data  & Mt-Wd & Fr-Ki & DMC & Adro & Tr-fi &WA \\ \hline
DINOv2~\cite{dinov2}          & LVD-142M                         &  77.9         &41.2          &59.4        &50.7    &69.0 &59.6        \\

CLIP~\cite{clip}         &    Image-text pairs                    & 75.5 & 39.8        &   52.2        & \textcolor{red}{\textbf{51.3}}       &  57.7          &55.6\\ 
R3M~\cite{r3m}         &    Ego                    & 81.3 &  30.6       &   52.2        & 46.7     &  64.7         &54.9 \\


\rowcolor{blue!5}VC-1~\cite{vc1}           &Ego\textsuperscript{*}+MNI                        & 88.8 & 38.4        & 60.9          &  46.0      &   70.5      &61.8 \\
\rowcolor{blue!5}MAE~\cite{mae}          & Ego                 &    85.1 & 36.7      &  59.2         &43.4        & \textcolor{red}{\textbf{70.6}}        &59.6  \\
\rowcolor{blue!5}STP         & Ego                 &    92.0 &  40.9    &  \textcolor{red}{\textbf{62.1}}         &  48.0      & 69.3        &63.7 \\ 

\rowcolor{blue!5}STP         & Ego+I                 & \textcolor{red}{\textbf{94.1}} & \textcolor{red}{\textbf{42.5}}     & 61.6      &  47.3      & 66.7        &\textcolor{red}{\textbf{64.2}}  \\ \hline
\rowcolor{blue!5}MAE (Post PT)           & Ego+Demo                  &   93.6 & 46.9      &81.1           &58.0        & 76.8         &72.5 \\
\rowcolor{blue!5}STP (Post PT)         & Ego+Demo                 & \textcolor{blue}{\textbf{97.3}} &\textcolor{blue}{\textbf{53.6}}      &   \textcolor{blue}{\textbf{82.8}}    &   \textcolor{blue}{\textbf{63.3}}     & \textcolor{blue}{\textbf{78.0}}          &\textcolor{blue}{\textbf{76.4}}\\ 
STP (E2E FT)         & Ego                 &87.2 & 52.4     &  55.2         &  40.0      &  70.4       &62.9\\ 
STP         & Demo                 &  70.3 & 30.4     & 52.5        & 38.0      &  70.8        &51.8 \\ 
\hline

\rowcolor{blue!5}
VC1-L/16 (Post PT)           & Ego+Demo                  &  95.7 & 54.7      &83.5           &66.0        &  \textbf{77.6}         &76.7 \\
\rowcolor{blue!5}
STP-L/16 (Post PT)         & Ego+Demo                 & \textbf{97.3} &\textbf{57.4}      &   \textbf{85.0}    &   \textbf{70.0}     & 75.4          &\textbf{78.4}\\

\hline
\end{tabular}
\caption{Performance on single-task simulation benchmarks. * denotes that VC-1 samples images form full Ego4D. Me, Fra, DMC, Adr, Tri, and WA represent MetaWorld, Franka-Kitchen, DMControl, Adroit, Trifinger, and weight average. We run the comparison between STP and MAE three times and report the mean and standard deviation in our appendix to demonstrate the stability of our improvments.}
 \label{sota}
\end{table*}

\begin{table*}[htb]
\centering
\begin{tabular}{ccccccc|cc}
\hline
          & DINOv2 & CLIP & R3M  & VC-1 & MAE  & \cellcolor{blue!5}{STP}  & VC1-L/16 & \cellcolor{blue!5}{STP-L/16} \\ \hline
LIBERO-LONG & 24.3±0.8      & 28.7±1.4 & 36.3±2.0 & 35.2±3.2 & 37.3±1.2 & \cellcolor{blue!5}{\textcolor{red}{\textbf{41.3±0.3}}} &    34.0±0.5  &  \cellcolor{blue!5}{\textbf{36.7±1.3} }    \\
RLbench  &  48.6±0.2     & 39.8±2.2  &45.9±0.8  &46.4±0.9  &45.2±0.2 & \cellcolor{blue!5}{\textcolor{red}{\textbf{51.3±1.4}}} & 57.5±0.4     & \cellcolor{blue!5}{\textbf{58.6±0.9}}    \\ \hline

\end{tabular}
\vspace{-1mm}
\caption{Performance comparations on multi-task simulation benchmarks.}
 \label{sota2}
  \vspace{-2.0mm}
\end{table*}

\subsection{Ablation on Downstream Simulation Tasks}

In this section, we perform extensive ablation studies in single-task simulation benchmarks, which further demonstrate the effectiveness of our joint spatial and temporal prediction, as well as temporal prediction condition design. In addition, we also study the influence of temporal decoder architecture design and future frame sampling strategy.

\noindent \textbf{Current frame masking.} 
The design of the current frame masking is crucial. On one hand, similar to MAE~\cite{mae}, masking some patches and predicting the missing parts can effectively promote the learning of image content features. On the other hand, the visible patches of the current frame need to interact with the condition to predict the future frame. Specifically, we mask the current frame at masking rates of 75\%, 50\%, and 0\%, respectively, and optionally predict the missing parts through the spatial decoder. The results are shown in Tab.~\ref{ablation} (a). From results, we see that the masking ratio of 75\% and performing spatial prediction still lead to the best performance. This demonstrates the importance of retaining MAE~\cite{mae} for content features learning, especially for low-level manipulation in Meta-World, while a current frame with a high masking ratio (75\%) is sufficient to interact with other conditions to predict the future frame.

\begin{table}[htbp]
\scalebox{0.70}{
\begin{tabular}{ccccccc}
\hline
                        & Push               & Pour                 & Fold                  & Tissue                 & Pass                  & WA                  \\ \hline
\multirow{2}{*}{DINOV2} &  \multicolumn{1}{l}{33.3±7.6}                     &  \multicolumn{1}{l}{50.0±8.7}                     &           \multicolumn{1}{l}{31.7±2.9}            &     \multicolumn{1}{l}{30.0±5.0}                   &    \multicolumn{1}{l}{26.7±2.9}                &   34.3             \\
                        &      \multicolumn{1}{l}{8.3±2.9}                &   \multicolumn{1}{l}{48.3±2.9}                   &         \multicolumn{1}{l}{26.7±5.8}             &   \multicolumn{1}{l}{0.0±0.0}                      &      \multicolumn{1}{l}{18.3±2.9}                &   20.3                  \\ \hline
\multirow{2}{*}{CLIP}   &        \multicolumn{1}{l}{31.8±5.8}              &  \multicolumn{1}{l}{51.7±5.8}                       &          \multicolumn{1}{l}{33.3±7.6}             &        \multicolumn{1}{l}{11.7±2.9}               &               \multicolumn{1}{l}{18.3±2.9}         &             29.4         \\
                        &   \multicolumn{1}{l}{16.7±2.9}                   &    \multicolumn{1}{l}{50.0±5.0}                   &  \multicolumn{1}{l}{31.7±2.9}                     & \multicolumn{1}{l}{0.0±0.0}                      &      \multicolumn{1}{l}{16.7±5.8}                  &   23.0                   \\ \hline
\multirow{2}{*}{R3M}    &  \multicolumn{1}{l}{31.7±5.8}                    &  \multicolumn{1}{l}{60.0±5.0}                     &  \multicolumn{1}{l}{35.0±5.0}                    & \multicolumn{1}{l}{38.3±10.4}                      &                  \multicolumn{1}{l}{30.0±5.0}      &  39.0                    \\
                        &   \multicolumn{1}{l}{25.0±5.0}                   &        \multicolumn{1}{l}{55.0±5.0}                & \multicolumn{1}{l}{35.0±5.0}                     &  \multicolumn{1}{l}{28.3±7.6}                     &       \multicolumn{1}{l}{25.0±5.0}                 &      33.7               \\ \hline

\multirow{2}{*}{MAE}    & \multicolumn{1}{l}{41.7±2.9} & \multicolumn{1}{l}{35.0±5.0} & \multicolumn{1}{l}{43.3±2.9} & \multicolumn{1}{l}{38.3±5.8} & \multicolumn{1}{l}{45.0±5.0} & 40.7\\
                        & \multicolumn{1}{l}{43.3±10.4} & \multicolumn{1}{l}{33.0±2.9} & \multicolumn{1}{l}{41.7±2.9} & \multicolumn{1}{l}{41.7±2.9} & \multicolumn{1}{l}{16.7±5.8} & 35.3\\ \hline
\multirow{2}{*}{VC-1}   & \multicolumn{1}{l}{30.0±5.0} & \multicolumn{1}{l}{38.3±12.6} & \multicolumn{1}{l}{48.3±5.8} & \multicolumn{1}{l}{45.0±5.0} & \multicolumn{1}{l}{48.3±7.6} &  42.0\\
                        & \multicolumn{1}{l}{31.7±2.9} & \multicolumn{1}{l}{20.0±10.0} & \multicolumn{1}{l}{45.0±0.0} & \multicolumn{1}{l}{45.0±5.0} & \multicolumn{1}{l}{41.7±2.9} & 36.7 \\ \hline
\multirow{2}{*}{STP}    & \multicolumn{1}{l}{56.7±5.8} & \multicolumn{1}{l}{61.7±5.8} & \multicolumn{1}{l}{43.3±2.9} & \multicolumn{1}{l}{43.3±5.8} &\multicolumn{1}{l}{41.7±2.9}   & \textbf{49.3}  \\
                        & \multicolumn{1}{l}{51.7±7.6} & \multicolumn{1}{l}{53.3±7.6} & \multicolumn{1}{l}{45.0±0.0} & \multicolumn{1}{l}{43.3±2.9} & \multicolumn{1}{l}{38.3±7.6} &\textbf{46.3} \\ \hline

\end{tabular}}
 \caption{\textbf{Performance comparations on real-world tasks.} For each task, the upper part indicates the result of the standard environment, and the lower part indicates the result of generalizing to a unseen environment with more distractors.}
 \label{sota33}
 \vspace{-6mm}
\end{table}

\begin{table*}[htbp]
\centering
\label{ablation}

\begin{minipage}{0.49\textwidth}
\centering
\resizebox{\textwidth}{!}{%
\begin{tabular}{c|c|c|c|c|c|c|c}
\toprule
$\rho^c$ & Predict & Me & Fra & DMC & Adr & Tri & WA \\
\cmidrule(r){1-8}
\rowcolor{blue!10}75\% &\checkmark &\textbf{92.0} & \textbf{40.9} & \textbf{62.1} & \textbf{48.0} & \textbf{69.3} & \textbf{63.7} \\
75\% & & 84.5 & 34.7 & 55.4 & 43.3 & 65.3 & 57.4 \\
50\% &\checkmark & 82.1 & 36.0 & 60.3 & \textbf{48.0} & 66.8 & 59.0 \\
0\% & & 79.2 & 39.7 & 54.8 & 44.0 & 63.1 & 57.0 \\
\bottomrule
\end{tabular}%
}
\caption*{(a) Current Frame Masking and Spatial Prediction.}
\end{minipage}
\hfill
\begin{minipage}{0.44\textwidth}
\centering
\resizebox{\textwidth}{!}{%
\begin{tabular}{c|c|c|c|c|c|c}
\toprule
Condition & Me & Fra & DMC & Adr & Tri & WA \\
\midrule
L-E & 82.1 & 30.7 & 55.5 & 42.0 & 63.8 & 55.4 \\
\rowcolor{blue!10}95\% & \textbf{92.0} & 40.9 & 62.1 & \textbf{48.0} & 69.3 & \textbf{63.7} \\
90\% & 91.2 & \textbf{42.5} & 62.8 & 44.7 & 65.9 & 63.4 \\
L-E + 95\% & 91.0 & 37.7 & \textbf{64.1} & 46.7 & \textbf{70.8} & 63.1 \\
L-D + 95\% & 88.0 & 34.3 & 62.6 & 46.7 & 69.3 & 60.9 \\
\bottomrule
\end{tabular}%
}
\caption*{(b) Temporal Prediction Condition Design.}
\end{minipage}



\begin{minipage}{0.49\textwidth}
\centering
\resizebox{\textwidth}{!}{%
\begin{tabular}{c|c|c|c|c|c|c}
\toprule
Decoder & Me & Fra & DMC & Adr & Tri & WA \\
\midrule
8 joint-self & 87.7 & 36.9 & 55.7 & 46.0 & \textbf{71.3} & 59.8 \\
12 joint-self & 88.5 & 35.0 & 55.7 & 46.0 & 67.0 & 59.1 \\
\rowcolor{blue!10}8 self-cross & \textbf{92.0} & \textbf{40.9} & \textbf{62.1} & \textbf{48.0} & 69.3 & \textbf{63.7} \\
\bottomrule
\end{tabular}%
}
\caption*{(c) Temporal Decoder Architecture Design.}
\end{minipage}
\hfill
\begin{minipage}{0.49\textwidth}
\centering
\resizebox{\textwidth}{!}{%
\begin{tabular}{c|c|c|c|c|c|c}
\toprule
Frame interval & Me & Fra & DMC & Adr & Tri & WA \\
\midrule
8 & 89.6 & 39.9 & 58.4 & 46.0 & 67.0 & 61.3 \\
\rowcolor{blue!10}16 & 92.0 & 40.9 & \textbf{62.1} & \textbf{48.0} & \textbf{69.3} & \textbf{63.7} \\
24 & 89.1 & \textbf{41.1} & 61.5 & 46.0 & 68.1 & 62.5 \\
8, 24 & \textbf{92.3} & 37.1 & 57.3 & 42.0 & 68.4 & 60.8 \\
\bottomrule
\end{tabular}%
}
\caption*{(d) Frame Sampling Strategy.}
\end{minipage}
\vspace{-2.5mm}
\caption{The ablation experiment results. Me, Fra, DMC, Adr, Tri, and WA respectively represent MetaWorld, Franka-Kitchen, DMControl, Adroit, Trifinger, and weight average.}
\label{ablation}
\vspace{-4.0mm}
\end{table*}

\noindent \textbf{Temporal prediction condition design.} Subsequently, we discuss the influence of temporal prediction condition design. We implicitly model motion in actionless video data by predicting the pixels of the future frame. A direct and simple idea is to use language narration as a condition. The text tokens can be flexibly utilized as inputs to ViT~\cite{vit}, forming a multimodal encoder. Language narration provides a high-level behavior description, but lacks low-level visual dynamic priors for pixel-level prediction. However, leaking part of the future frame can effectively provide these priors.  In order to explore how to construct a more meaningful temporal prediction proxy task, we compare the following schemes: (1) only language narration, (2) masking 95\% of the future frame, (3) masking 90\% of the future frame, (4) masking 95\% of the future frame and language narration, and (5) masking 95\% of the future frame and language narration, but the language is added in the temporal decoder, instead of being fused with the visible image patches in the multimodal encoder. We tokenize all language narration by pre-trained DistilBERT~\cite{distilbert}. The results are shown in Tab.~\ref{ablation} (b). From results, we see that using only language as a prediction condition leads to a significant decline in performance, while leaking a small amount of future frame (masking 95\%) in the temporal decoder can achieve competitive results. As for joint conditions of language and future frame with 95\% masking ratio, adding language in the encoder is better than in the decoder. Additionally, adding language performs better on DMControl (64.1 vs. 62.1) and Trifinger (70.8 vs. 69.3), while not adding language performs better on Meta-World (92.0 vs. 91.0), Franka-Kitchen (40.9 vs. 37.7) and Adroit (48.0 vs. 46.7). We speculate the reasons for language hurts performance are as follows: (\romannumeral 1) The input gap (multi-modal and single-modal) between upstream and downstream; (\romannumeral 2) Extra language in ViT may result in the loss of some fine-grained information capture. Furthermore, the latter does not require language supervision, and can provide a more scalable and promising self-supervised solution.



\noindent \textbf{Temporal decoder design.} 
We also investigate the impact of the temporal decoder design. Specifically, we consider two types of decoder blocks. One is the joint-self architecture, as shown in Fig.~\ref{decord} (a), and similar interaction architecture are adopted in ~\cite{m3ae,comae}. The other is the self-cross architecture, as shown in Fig.~\ref{decord} (b), and similar interaction architecture are adopted in ~\cite{multimae,mixformer_j,siammae}. We consider the following settings: (1) 8 joint-self decoder blocks, (2) 12 joint-self decoder blocks, (3) 8 self-cross decoder blocks. Among them, setting (2) and (3) have similar amounts of parameters for a fairer comparison. The results are shown in Tab.~\ref{ablation} (c). The results demonstrate the importance of that the current frame representation cannot be updated in temporal decoder and cannot see the future frame feature.

\noindent \textbf{Frame sampling strategy.} 
Finally, we investigate the impact of the sampling strategy between the current frame and future frame. The difficulty of temporal prediction is directly proportional to the frame interval values. We establish four settings where we fix the sampling intervals at 8, 16, and 24 respectively, and for the fourth setting, we randomly select an interval within the range of [8, 24]. The results are shown in Tab.~\ref{ablation} (d). The results show that an interval of 16 achieves the best balance for building temporal prediction proxy task.

\begin{figure}[htbp]
\centering
\includegraphics[width=0.45\textwidth]{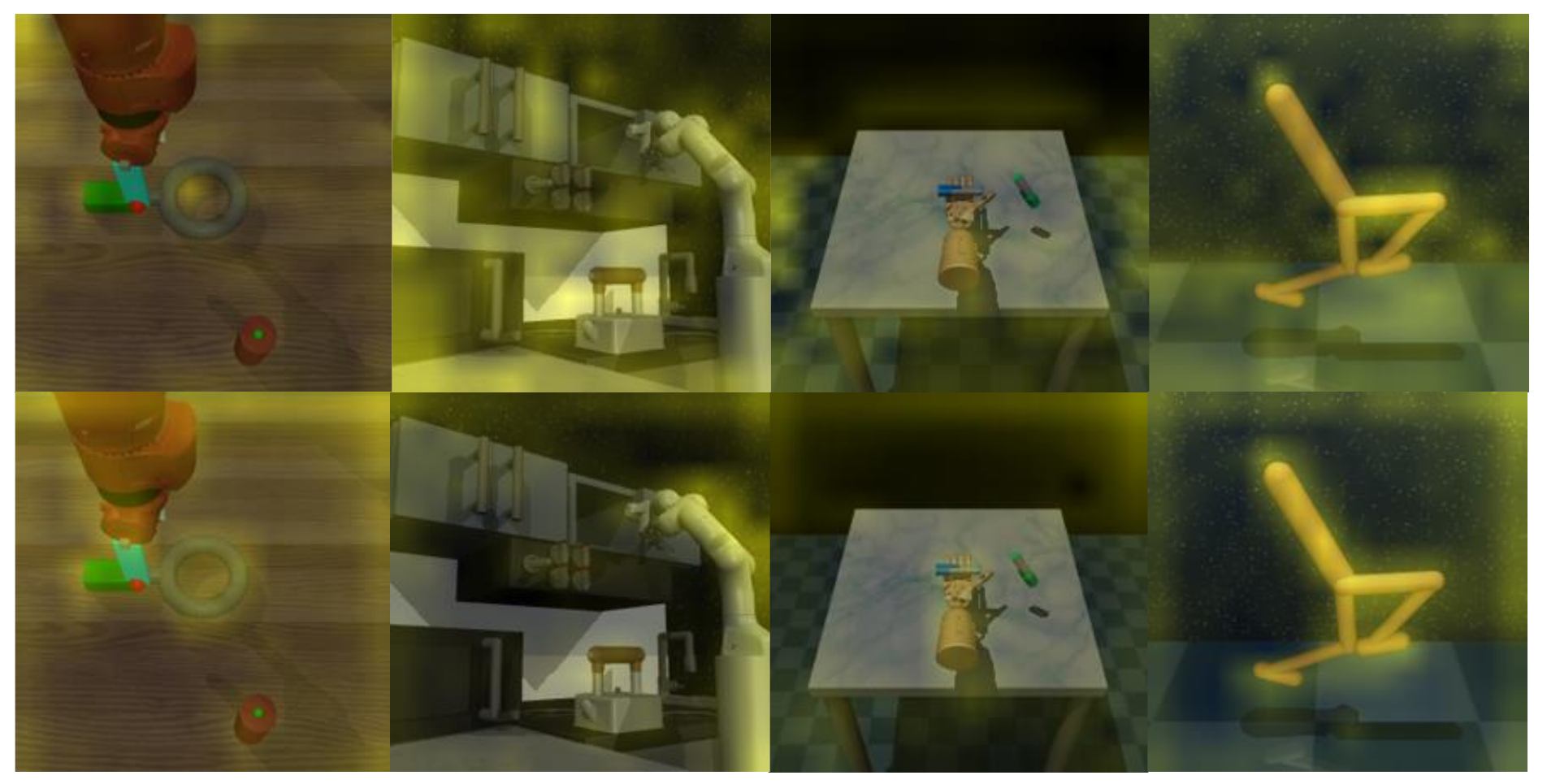} 
\label{viss}
\vspace{-4.0mm}
\caption{\textbf{Attention Visualization.} We use the [CLS] token as query, average the attention of all heads at the last layer of the frozen ViT encoder, where the size of the attention value is directly proportional to the intensity of the yellow light. \textbf{Top:} MAE pre-training. \textbf{Bottom:} STP pre-training. } 
\label{viss}
\end{figure}

\vspace{-4.0mm}

\vspace{-2mm}


\section{Conclusion}
In this work, we propose STP, a simple, efficient and effective self-supervised image representation pre-training framework for robotic motor control. Our STP jointly performs spatiotemporal prediction on large-scale videos within a multi-task learning manner. Our STP captures content features by predicting the invisible areas within the masked current frame, and simultaneously captures motion features by using a future frame with an extremely high masking ratio as a condition to predict the invisible areas within that future frame. We carry out extensive simulation and real-world experiments to demonstrate the effectiveness and generalization capabilities of STP, especially in generalizing to unseen environments with more distractors. Furthermore, as for pre-training data, we also prove that extending STP to hybrid pre-training and post-pre-training could further unleash its generality and data efficiency.

\appendix

\renewcommand\thesection{\Alph{section}} 

\section{Appendix}

\subsection{Limitations and Discussion}

Our STP jointly conducts spatial prediction and temporal prediction by asymmetric masking and decoupled dual decoders for self-superviesd robotic motor control image representation pre-training. Although STP has demonstrated superior performance on extensive simulation and real-world experiments, there still remain some challenges and future works. Specifically, on the one hand, we pre-train a plain and general image ViT, without designing better transfer modules tailored for specific downstream policy models and application scenarios; we leave this for future work. On the other hand, our method is self-supervised, and an interesting avenue for future exploration is to combine it with supervised pre-training methods, such as language, depth, action, bounding box, and additional teachers, to build a more powerful foundation model. Finally, regarding our pre-training methods, exploring predictive targets beyond pixel space and more effective sampling and masking strategies also present intriguing research directions.

\subsection{The stability of our simulation single-task experiments}

As for our simulation single-task experiments, the number of total evaluation episodes = the number of tasks × the number of BC seeds × the number of evaluation seeds × the number of camera views. Therefore, as shown in Tab.~\ref{tongji}, the total number of our evaluation episodes for each representation is $5\times3\times25\times1+5\times3\times50\times2 + 5\times3\times25\times1 + 2\times3\times25\times1 + 2\times1\times25\times1 = \mathbf{2450}$. Therefore, the scale of our simulation single-task evaluation is \textbf{enormous}.

For the same BC seed, our policy training is fully reproducible, with the only uncertainty being the slight difference that still exist in MuJoCo rendering even with the same policy and evaluation seed. Therefore, we further rerun our paper's STP-B and MAE-B baseline twice, obtaining three results and their mean and standard deviation, as shown in Tab.~\ref{rerun}. The results show that STP-B outperforms MAE-B by 5.2 (63.9 vs. 58.7), which proves that our improvements are \textbf{significant and stable}. Additionally, the rendering in Trifinger is not subject to randomness, hence the results are fully reproducible.

\begin{table*}[h]
\centering
\resizebox{\textwidth}{!}{
\begin{tabular}{ccccc}
\toprule
 & \textbf{The number of BC seeds} & \textbf{The number of evaluation seeds} & \textbf{The number of camera views} &\textbf{The number of tasks}\\
\midrule
Meta-World & 3 & 25 & 1 &5 \\
Franka-Kitchen & 3 & 50 & 2  &5\\
DMControl & 3 & 25 & 1 &5\\
Adroit & 3 & 25 & 1 &5\\
Trifinger & 1 & 25 & 1 &2\\
\bottomrule
\end{tabular}
}
\caption{The evaluation details on simulation single-task benchmarks.}
\label{tongji}
\end{table*}

\begin{table*}[h]
\centering
\begin{tabular}{cccc}
\toprule
 & \textbf{BC seeds × evaluation seeds × run times} & \textbf{STP-B} & \textbf{MAE-B} \\
\midrule
Meta-World & 25 × 3 × 3 & 94.1, 93.6, 94.1, \textbf{93.9±0.2} & 85.1, 84.8, 84.3, \textbf{84.7±0.3} \\
Franka-Kitchen & 50 × 3 × 3 & 42.5, 43.5, 43.8, \textbf{43.3±0.6} & 36.7, 37.9, 37.1, \textbf{37.2±0.5} \\
DMControl & 25 × 3 × 3 & 61.6, 60.3, 60.7, \textbf{60.9±0.5} & 59.2, 60.3, 60.3, \textbf{59.9±0.5} \\
Adroit & 25 × 3 × 3 & 47.3, 48.7, 48.0, \textbf{48.0±0.6} & 43.4, 45.3, 44.7, \textbf{44.5±0.8} \\
WA & 25 × 3 × 3 & 63.9, 63.8, 64.1,\textbf{ 63.9±0.3} & 58.3, 59.1, 58.7, \textbf{58.7±0.3} \\
\bottomrule
\end{tabular}
\caption{The additional comparison results. }
\label{rerun}
\end{table*}

\subsection{The additional comparison results}
In this section, we further add three additional methods in the simulation single-task evaluation, which are the publicly available LIV~\cite{liv} (ResNet-50), the publicly available MPI (ViT-base)~\cite{mpi}, and the VideoMAE (4-frame) trained based on EgoClip. Among them, for MPI, we remove the additional multi-frame causality modeling, multimodal token aggregator and multiheaded attention pooling, also using the [CLS] token for a fair comparison. The results, as shown in Tab.~\ref{add}, still demonstrate a significant advantage for STP. We believe that the poorer performance of VideoMAE is due to the gap in temporal interaction between upstream and diverse downstream environments, which lead to a significant risk of cumulative error in the paradigm of imitation learning.

\begin{table*}[htbp]
\centering
\label{tab:my_label}
\begin{tabular}{@{}ccccccc@{}}
\toprule
& Meta-World & Franka-Kitchen & DMControl & Adroit & Trifinger & WA \\
\midrule
STP (EgoClip) & 92.0 & 40.9 & 62.1 & 48.0 & 69.3 & 63.7 \\
VideoMAE (EgoClip) & 68.5 & 30.5 & 53.9 & 47.3 & 70.5 & 52.6 \\
LIV (ResNet-50) &81.3 &37.3 &54.0 &52.0 &68.3 &58.1 \\
MPI (ViT-base) &82.1 &38.4 &55.7 &49.3 &67.7 &58.7 \\
\bottomrule
\end{tabular}
\caption{Comparison of STP (EgoClip) and MAE-ST (EgoClip) on various benchmarks.}
\label{add}
\end{table*}

\subsection{Pre-training Details}

\begin{figure*}[htbp]
\centering
\includegraphics[width=0.9\textwidth]{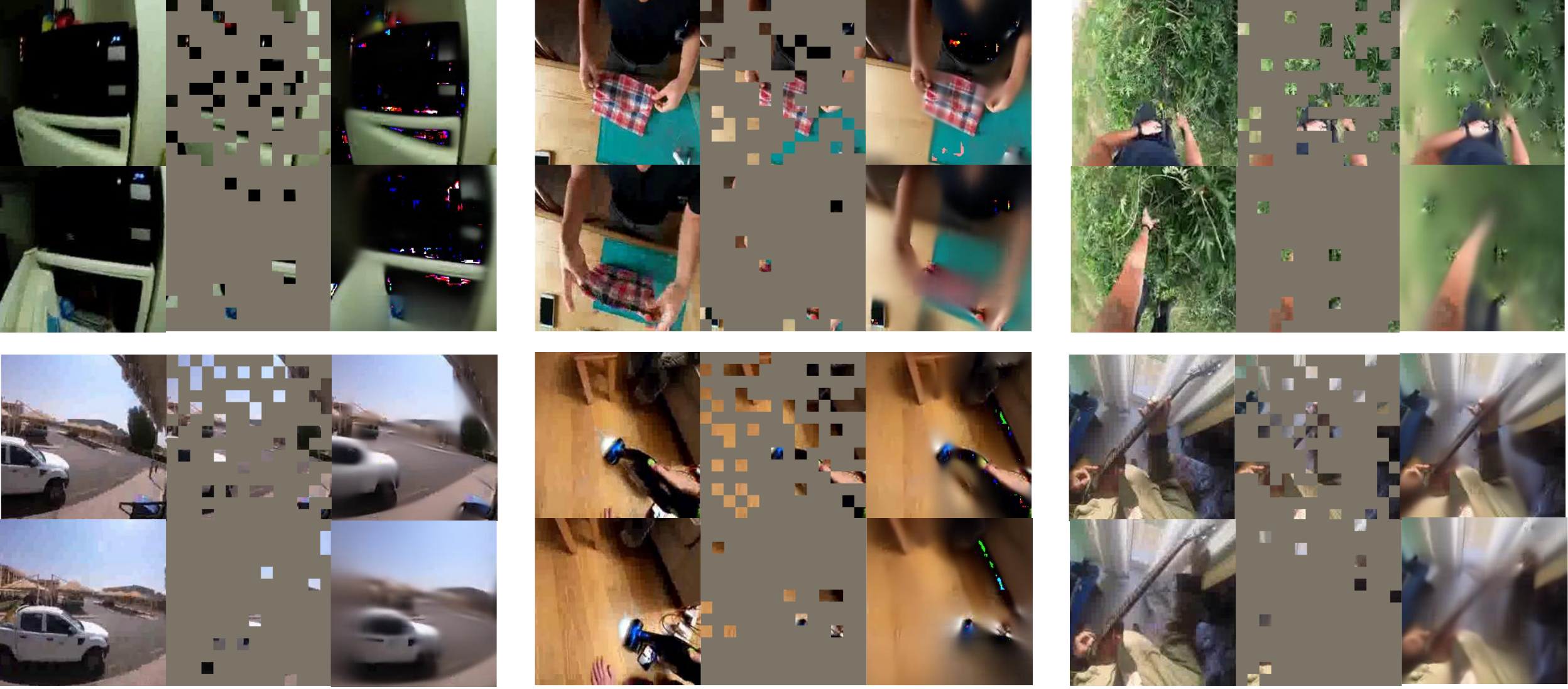} 
\label{method}
    \caption{Some examples of our STP prediction result on Ego4D videos. For each six tuple,
we show the ground-truth (left), masked frames (middle), STP prediciton results (right), current frames (top), and future frames (bottom). We simply overlay the
output with the visible patches to improve visual quality.}
\label{vis}
\end{figure*}

In this section, we describe the details of our STP pre-training. Specifically, we list some key training and architectural hyperparameters of STP in Tab.~\ref{hyperparams}. In addition, as for our MAE~\cite{mae} baseline, we mainly follow the publicly available code of MAE\footnote{\url{https://github.com/facebookresearch/mae}}. Additionally, we train MAE and STP using the same hyperparameters, data and number of epochs to ensure that the comparison between them is completely \textbf{fair}. Finally, we also provide some STP prediction results in Fig.~\ref{vis}.

\begin{table*}[hbp]
    \centering
    \setlength{\tabcolsep}{10pt}
    \begin{tabular}{ll}
        \toprule
        \textbf{Hyperparameter} &  \textbf{Value}\\\hline
        \multicolumn{2}{c}{\textit{STP Pre-training}}\\\hline
        optimizer & AdamW~\cite{adam} \\
        base learning rate & 0.00015 \\
        weight decay & 0.05 \\
        optimizer momentum & $\beta_1, \beta_2=0.9, 0.95$\\
        effective batch size & 4096 \\
        learning rate schedule & cosine decay \\
        total epochs & 50\\
        warmup epochs & 5 \\
        augmentation & {RandomResizedCrop (0.8, 1)} \\
        \hline
        \multicolumn{2}{c}{\textit{Encoder ViT-base Architecture}}\\
        \hline
         patch size & 16\\
        \#layers & 12\\
        \#MHSA heads & 12\\
        hidden dim & 768\\
        positional embedding & sin-cos initialization and fix\\
        \hline
        \multicolumn{2}{c}{\textit{Dual Decoder ViT-base Architecture}}\\
        \hline
        \#layers & 8\\
        \#MHSA heads & 16\\
        hidden dim & 512\\
        positional embedding & sin-cos initialization and fix\\
        \bottomrule
    \end{tabular}
    \caption{Training and architectural hyperparameters for STP pre-training.}
     \label{hyperparams}
\end{table*}

\subsection{The influence of the loss weight ratio between temporal prediction and spatial prediction}

\begin{figure*}[htbp]
\centering
\includegraphics[width=0.85\textwidth]{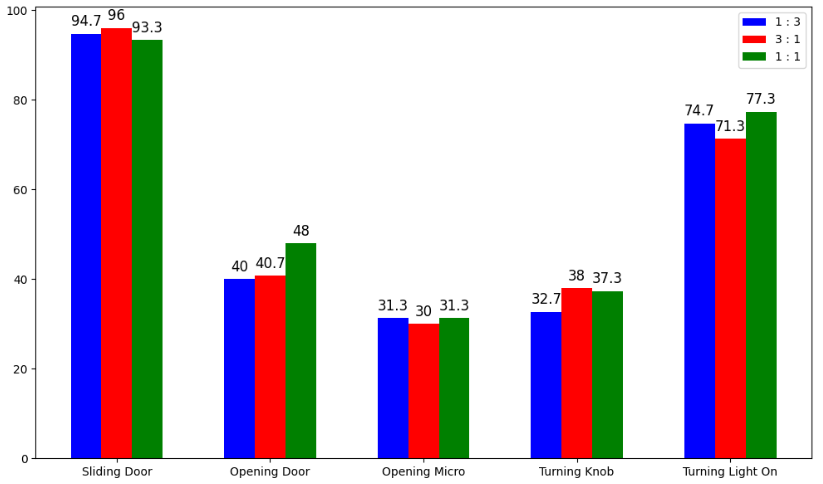} 
\label{loss}
\caption{The results of different loss weight ratios between temporal prediction and spatial prediction.}
\label{franka}
\vspace{-4mm}
\end{figure*}

In this section, we further explore the influence of the loss weight ratio between temporal prediction and spatial prediction. Specifically, taking five tasks from Franka-Kitchen as examples, we load the pre-trained STP and perform post-pre-training with three different loss weight ratios (temporal to spatial). The results, as shown in Fig.~\ref{franka}, are 54.7, 55.2, and 57.4 for the average results of the ratios 3:1, 1:3, and 1:1, respectively. The results indicate that due to the different attributes of the tasks, the trends are not consistent. However, overall, the 1:1 ratio achieves the best balance and results. We chose it as a universal setting.

\subsection{The comparison on reinforcement learning setup}

We also compare our STP-B with the publicly available and the most advanced VC1-B in the setting of reinforcement learning. Specifically, we select the Panda-Door and Panda-TwoArmPegInHole tasks from the Robosuite~\cite{robosuite} simulation environment for reinforcement learning evaluation. We employ DrQ-v2 as our RL algorithm and compare the results of VC-1 (ViT-B) and our STP (ViT-B) as frozen visual representations. Due to the large fluctuations in success rate, we report the maximum reward value under 200,000 steps in Tab.~\ref{rlrl}, and the results verify the effectiveness of our STP within the RL framework.

\begin{table*}[h]
\centering
\begin{tabular}{@{}ccc@{}}
\toprule
Task & STP & VC-1 \\
\midrule
Panda-Door & 95.6 & 88.8 \\
Panda-TwoArmPegInHole & 130.7 & 123.1 \\
\bottomrule
\end{tabular}
\caption{The comparison of STP and VC-1 on  reinforcement learning setup.}
\label{rlrl}
\end{table*}

\subsection{Simulation single-task evaluation environments details}

In this section, we first present further details of the STP post-pre-training on single-task simulation environments. Subsequently, we delineate the specific hyperparameters used in the behavior cloning policy training within these single-task simulation environments. Finally, we provide the comprehensive evaluation scheme for each simulation environment. 

In regards to the STP post-pre-training, we utilize data that aligns with the policy training, and the specific architecture hyperparameters correspond to those listed in Tab. \ref{hyperparams}. Depending on the specific demonstration data, we adjust the values of total epochs, warmup epochs, effective batch size, and the frame interval, as shown in Tab.~\ref{post}. 

As for policy training and evaluation schemes, we primarily refer to the publicly available code\footnote{\url{https://github.com/facebookresearch/eai-vc/tree/main/cortexbench}} and training data of VC-1~\cite{vc1} for Metaworld~\cite{metaworld}, DMControl~\cite{dmc}, Adroit~\cite{adroit} and Trifinger~\cite{trifinger}. Similarly, for Franka-Kitchen~\cite{franka}, we follow the public code\footnote{\url{https://github.com/facebookresearch/r3m/tree/eval/evaluation}} and training data of R3M~\cite{r3m}. Specifically, the policy training hyperparameters and evaluation schemes are shown in Tab.~\ref{bc} and Tab.~\ref{eval}, respectively. About policy training, we completely follow the setting of prior works~\cite{r3m,vc1} when freezing the encoder; when performing end-to-end fine-tuning, we make appropriate adjustments to the batch size and learning rate. About evaluation details, similar to prior works\cite{r3m, vc1}, we establish all evaluation details such as the number of expert demonstrations and test trajectories, environmental viewpoints, optimization hyperparameters, base seeds, history windows size, and the use of robot proprioceptive. In Tab.~\ref{eval}, the term `prop.' stands for whether proprioceptive information is used or not, and `history window size' signifies the number of frames received by the policy model at each step, with features between frames being fused through concatenation. `Number of trajectories' represents the quantity of trajectories evaluated. For tasks in Meta-World, Franka-Kitchen, Adroit, and Trifinger, we report the maximum success rate, whereas for tasks in DMControl, we report the maximum reward score, rescaling to be in the range of [0, 100] by dividing by 10. We report the average metric across tasks for each environment. In addition, it is worth noting that the metrics we report are the \textbf{average value across all base seeds and camera viewpoints}.


In addition, we emphasize that different random seeds primarily affect the rendering of the initial frame in the sampled trajectories. During evaluation, the seed value we provide serves as the base seed, and the trajectory sampling process is depicted in Algorithm~\ref{render}. \textbf{Therefore, the actual number of trajectories we evaluate is the number of trajectories multiplied by the number of base seeds.} For instance, for MetaWorld, we evaluate 25 × 3 = 75 trajectories, with random seeds for rendering being 100-124, 200-224, and 300-324. 

Finally, for Franka-Kitchen, we utilize MuJoCo210, while all other simulation environments are based on MuJoCo200. Our policy training and evaluation environments are conducted on Cuda 11.3, NVIDIA TITAN Xp GPUs, and OpenGL 3.1.

\begin{table*}[htbp]  
\centering
\begin{tabular}{c|c|c|c|c|c}  
\toprule   
                         & MetaWorld & \multicolumn{1}{l|}{Franka-Kitchen} & DMControl & Adroit & Trifinger \\ \hline  
total epochs             & 50        & 100                                 & 50        & 50     & 50        \\  
warmup epochs            & 5         & 5                                   & 5         & 5      & 5         \\  
effective batch size             & 1024        & 128                                 & 2048       & 1024    & 1024        \\  
number of demonstrations & 25        & 25                                  & 100       & 100    & 100       \\  
frame interval           & 4         & 4                                   & 4         & 4      & 16        \\   
\bottomrule 
\end{tabular}  
 \caption{STP post-pre-training hyperparameters on single-task simulation environments.}
  \label{post}
\end{table*}

\begin{table*}[htbp] 
\centering
\begin{tabular}{c|c|c|c|c|c|c}  
\hline  
                               & \multicolumn{1}{c|}{} & \multicolumn{1}{c|}{MetaWorld} & \multicolumn{1}{c|}{Franka-Kitchen} & \multicolumn{1}{c|}{DMControl} & \multicolumn{1}{c|}{Adroit} & \multicolumn{1}{c}{Trifinger} \\ \hline  
\multicolumn{1}{c|}{epochs}                         & \multicolumn{1}{c|}{} & 100       & 480                                 & 100       & 100    & 100 / 1000                   \\ \hline 
\multicolumn{1}{c|}{\multirow{2}{*}{batch size}}    & frozen                & 256       & 32                                  & 256       & 256    & \multicolumn{1}{c}{32} \\  
\multicolumn{1}{c|}{}                               & fine-tuning           & 64        & 32                                  & 64        & 64     & 16                    \\  \hline
\multicolumn{1}{c|}{\multirow{2}{*}{learning rate}} & frozen                &  0.001     & 0.001                                &  0.001      &  0.001   &0.0001               \\  
\multicolumn{1}{c|}{}                               & fine-tuning           & 0.00005      & 0.0001                                & 0.00005      & 0.00005    &  0.0001                    \\ \hline  
\end{tabular}  
 \caption{Policy training hyperparameters on sing-task simulation environments.}
  \label{bc}
\end{table*}

\begin{table*} [htbp]  
\centering  
\begin{tabular}{ccccccc}  
\toprule    
\begin{tabular}{@{}c@{}}{Benchmark}\\\end{tabular} &  
\begin{tabular}{@{}c@{}}{Observation}\\{Space}\end{tabular} &  
\begin{tabular}{@{}c@{}}{History}\\{Window Size}\end{tabular} &  
\begin{tabular}{@{}c@{}}{Camera}\\{ViewPoints}\end{tabular} &  
\begin{tabular}{@{}c@{}}{Base Seeds}\\\end{tabular} & 
\begin{tabular}{@{}c@{}}{Number of }\\{Trajectories}\end{tabular} \\  
\midrule   
Metaworld & RGB + prop.    & 3  &top\_cap2 &100, 200, 300 &25 \\
Franka-Kitchen & RGB + prop.   & 1 & left, right &123, 124, 125 & 50\\
DMControl & RGB    &  3 & 0 &100, 200, 300 &25\\   
Adroit & RGB + prop.   &1 & vil\_camera &100, 200, 300 &25 \\  
Trifinger & RGB + prop. & 1 & default &10 &25\\     
\bottomrule  
\end{tabular}  
\caption{Evaluation schemes on single-task simulation environments.}
\label{eval}  
\end{table*}

\begin{algorithm*}
\centering
\caption{Trajectories Sampling Pseudocode}
\label{render}
\definecolor{codeblue}{rgb}{0.28125,0.46875,0.8125}
\lstset{
    basicstyle=\fontsize{9pt}{9pt}\ttfamily\bfseries,
    commentstyle=\fontsize{9pt}{9pt}\color{codeblue},
    keywordstyle=
}
\begin{lstlisting}[language=python]
# num_traj: the number of evaluation trajectories
# base_seed: base seed for rollouts

# rollout to sample trajectories
    for ep in range(num_traj):
        seed = base_seed + ep
        env.set_seed(seed)
        o = env.reset()
\end{lstlisting}
\end{algorithm*}

\subsection{Simulation multi-task evaluation environments details}
In this section, we outline the details of our simulation multi-task evaluation environments details. Specifically, for LIBERO-LONG~\cite{libero}, we provide a detailed tasks list in Tab.~\ref{libero}. Given 10 tasks data, with 20 demonstrations per task, we train a multi-task transformer policy using frozen visual features. We train 25 epochs and directly use the model from the 25th epoch for evaluation, performing 20 trials for each task and repeating the run three times. For the RLBench~\cite{rlbench}, we also provide a list of tasks in Tab.~\ref{rlbench}. We train a multi-task RVT-2~\cite{rvt} policy using frozen visual representations. Each task utilizes 100 demonstrations and we train for only 1,2000 iterations (less than one-tenth of the original RVT-2 setup). Finally, we also perform 20 trials for each task evaluation and repeat the process three times.

\subsection{Real-World environments details}

\begin{figure*}[htbp]
\centering
\includegraphics[width=0.85\textwidth]{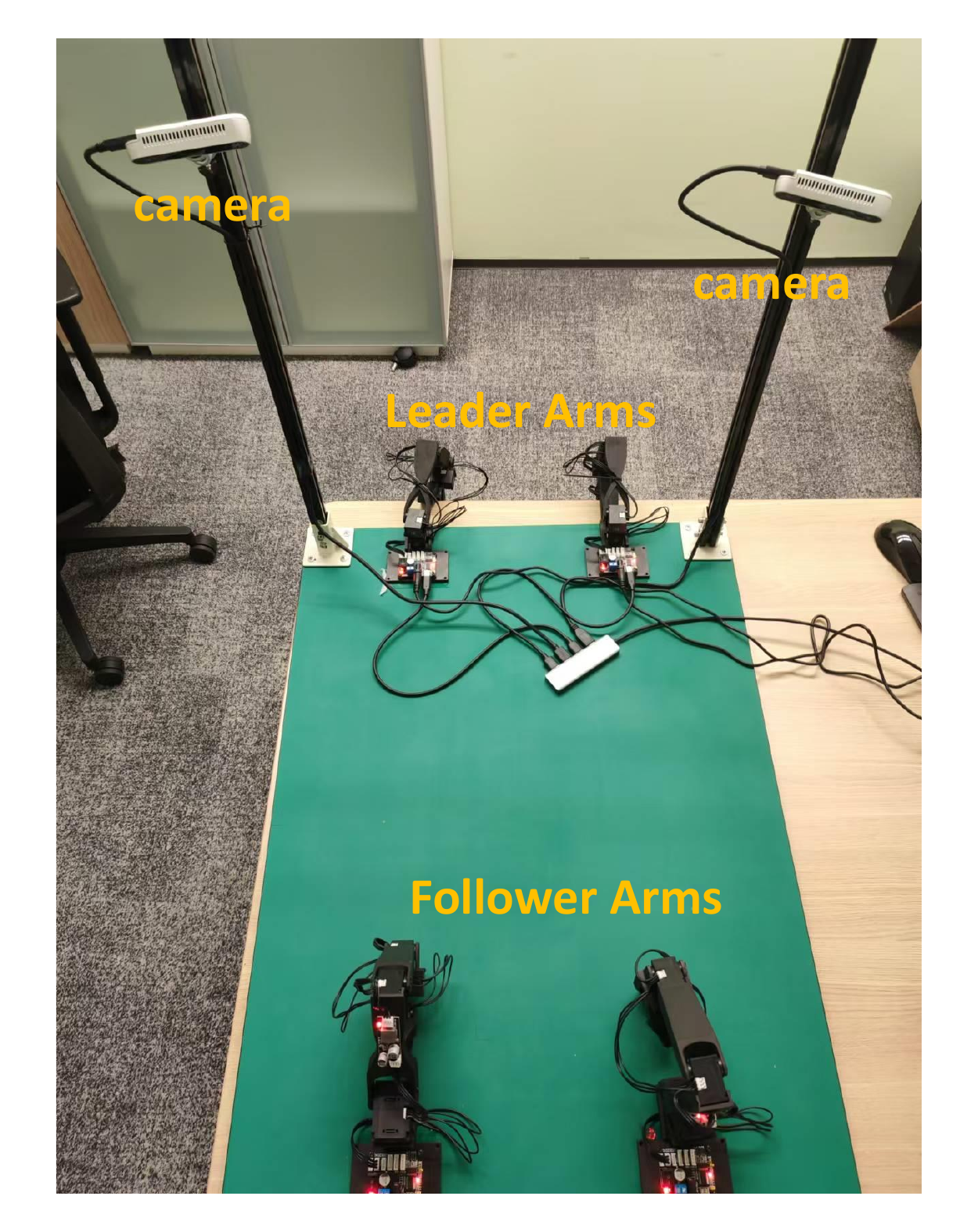} 
\caption{The real-world experiments hardware platform.}
\label{dualarm}
\vspace{-4mm}
\end{figure*}

In this section, we outline the details of our real-world setup and evaluation scheme. As depicted in Fig.~\ref{dualarm}, our real-world scenario includes two camera viewpoints: left and right. Specifically, we use two leader arms to perform teleoperation for follower arms data collection, where 50 demonstrations are collected for each task for ACT~\cite{act} policy training. For our real-world evaluation, we evaluate two setups: the standard environment and the unseen environment with more distractors, testing 20 trials for each setup and running each setup three times. Consequently, this results in a total of 120 ($2\times20\times3$) trials for each representation.

\begin{table*}[h]
\centering
\begin{tabular}{|c|}
\hline
turn on the stove and put the moka pot on it demo \\
\hline
put the black bowl in the bottom drawer of the cabinet and close it demo \\
\hline
put the yellow and white mug in the microwave and close it demo\\
\hline
put both moka pots on the stove demo \\
\hline
put both the alphabet soup and the cream cheese box in the basket demo \\
\hline
put both the alphabet soup and the tomato sauce in the basket demo \\
\hline
put both the cream cheese box and the butter in the basket demo \\
\hline
put the white mug on the left plate and put the yellow and white mug on the right plate demo \\
\hline
put the white mug on the plate and put the chocolate pudding to the right of the plate demo \\
\hline
pick up the book and place it in the back compartment of the caddy demo \\
\hline
\end{tabular}
\caption{The long-horizon LIBERO-LONG tasks for multi-task setup evaluation.}
\label{libero}
\end{table*}

\begin{table*}[h]
\centering
\begin{tabular}{|c|}
\hline
insert usb in computer \\
\hline
toilet seat down \\
\hline
sweep to dustpan\\
\hline
take plate off colored dish rack \\
\hline
open oven \\
\hline
stack blocks \\
\hline
beat the buzz \\
\hline
open wine bottle \\
\hline
open fridge \\
\hline
reach target \\
\hline
basketball in hoop \\
\hline
put rubbish in bin \\
\hline
meat on grill \\
\hline
take frame off hanger \\
\hline
hang frame on hanger \\
\hline
close laptop lid \\
\hline
take usb out of computer \\
\hline
change clock \\
\hline
close microwave \\
\hline
press switch \\
\hline
\end{tabular}
\caption{The RLbench tasks for multi-task setup evaluation.}
\label{rlbench}
\end{table*}

{
    \small
    \bibliographystyle{ieeenat_fullname}
    \bibliography{main}
}


\end{document}